\documentclass[letterpaper]{article} 
\usepackage{aaai24}  
\usepackage{times}  
\usepackage{helvet}  
\usepackage{courier}  
\usepackage[hyphens]{url}  
\usepackage{graphicx} 
\usepackage{natbib}  
\usepackage{caption} 
\usepackage{algorithm}
\usepackage{algorithmic}

\usepackage{newfloat}
\usepackage{listings}
\DeclareCaptionStyle{ruled}{labelfont=normalfont,labelsep=colon,strut=off} 
\floatstyle{ruled}
\newfloat{listing}{tb}{lst}{}
\floatname{listing}{Listing}
\usepackage{enumitem}
\usepackage{amsmath}
\usepackage{amsfonts}
\usepackage{subcaption}
\usepackage[table]{xcolor}
\usepackage{multirow}

\title{Detecting Statements in Text: A Domain-Agnostic Few-Shot Solution}
\author{
    Sandrine Chausson,
    Björn Ross
}
\affiliations{
    University of Edinburgh\\
    Informatics Forum \\
    10 Crichton St, Newington \\
    Edinburgh EH8 9AB \\

    sandrine.chausson@ed.ac.uk, b.ross@ed.ac.uk
}

\begin{document}

\maketitle

\begin{abstract}
Many tasks related to Computational Social Science and Web Content Analysis involve classifying pieces of text based on the claims they contain. State-of-the-art approaches usually involve fine-tuning models on large annotated datasets, which are costly to produce. In light of this, we propose and release a qualitative and versatile few-shot learning methodology as a common paradigm for any claim-based textual classification task. This methodology involves defining the classes as arbitrarily sophisticated taxonomies of claims, and using Natural Language Inference models to obtain the textual entailment between these and a corpus of interest. The performance of these models is then boosted by annotating a minimal sample of data points, dynamically sampled using the well-established statistical heuristic of Probabilistic Bisection. We illustrate this methodology in the context of three tasks: climate change contrarianism detection, topic/stance classification and depression-relates symptoms detection. This approach rivals traditional pre-train/fine-tune approaches while drastically reducing the need for data annotation.
\end{abstract}

\section{Introduction}\label{sec:intro}

\begin{figure}
\centering
  \includegraphics[width=8cm]{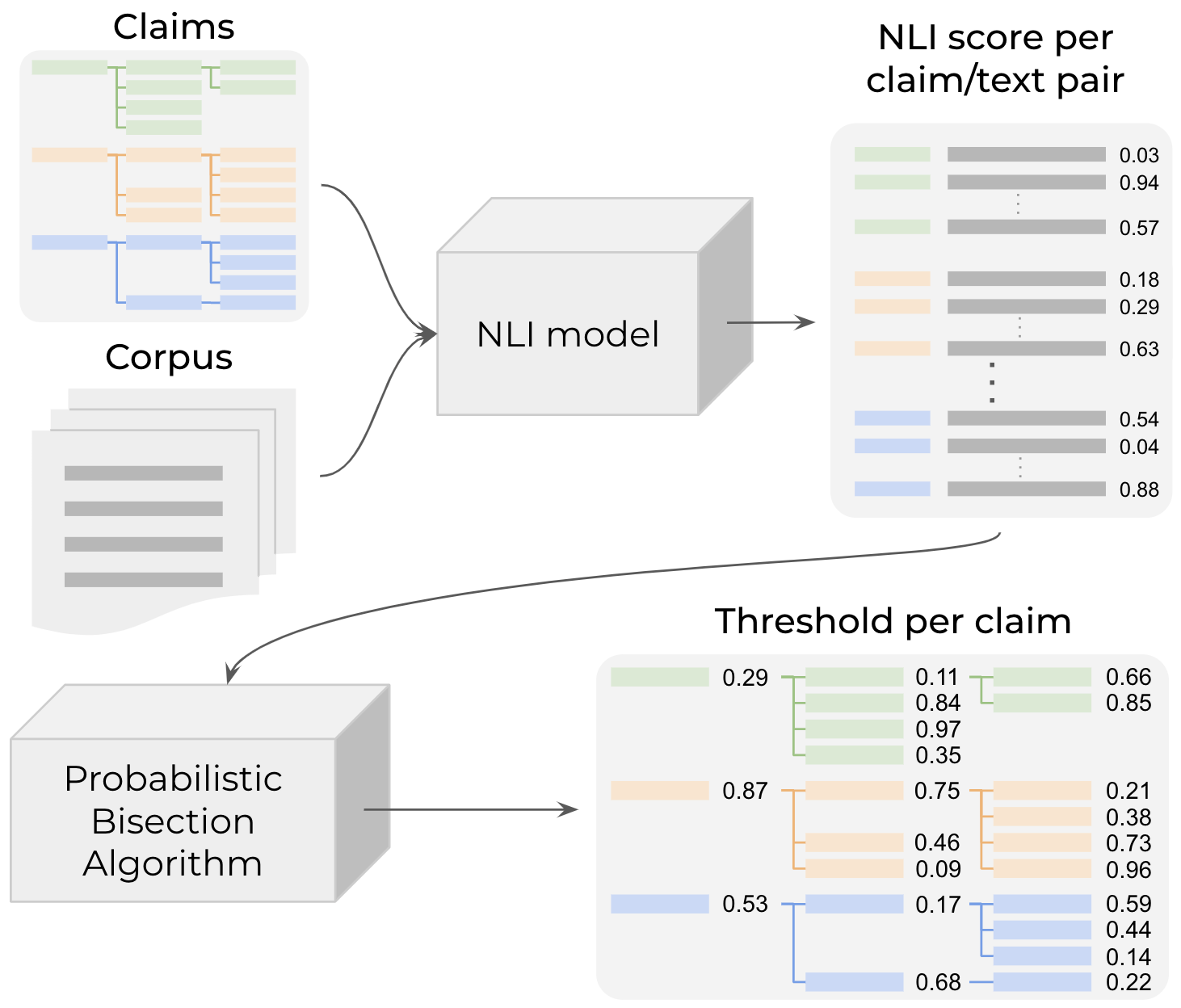}
  \caption{Overview of our proposed methodology. }
  \label{fig:teaser}
\end{figure}

Classifying text based on whether it contains certain statements is central to many Computational Social Science (CSS) and Web Content Analysis applications. These can be statements of support or opposition towards some target entity, event, policy or belief (\emph{stance} detection) \cite{mohammad2016semeval}. They can be untrue or misleading statements (\emph{misinformation} or \emph{disinformation} detection), for instance relating to Covid-19 \cite{khan2021detecting} or climate change \cite{coan2021computer}. Or they can be hateful statements (\emph{hate speech} detection) targeting individuals or groups based on some protected characteristic, such as race, gender or religious belief \cite{inproceedings}. Even \emph{fact-checking} \cite{zeng2021automated} can be construed as the detection of statements of interest in an ``evidence-providing" corpus, such as news articles or academic publications \cite{pomerleau2017fake}. 

To address these tasks, the field has produced a profusion of specialised models usually obtained by fine-tuning pre-trained language models on large manually annotated datasets. This is an expensive paradigm. Compiling and annotating datasets is costly in both time and money, and there are many examples of \emph{ad hoc} tasks where a social scientist may wish to define their own categories for which no manually annotated data is available. For example, one may wish to detect specific assertions in politician's Facebook and Twitter posts \cite{mcgregor2017personalization} or specific conspiracy theories in YouTube comments \cite{ROCHERT2022101866}. Even when datasets exist, these can quickly become obsolete or incomplete as the public discourse evolves in response to new events. This is particularly discomfiting in light of often extensive yet underutilised domain-specific theoretical resources. Moving away from training ``single-use" classifiers, towards a more general and data-frugal paradigm, particularly one that enables \emph{theory-grounded} classification, is therefore essential.

In this paper, we propose precisely such a shift. More concretely, we propose a domain-agnostic methodology for the detection of statements in natural text. The methodology leverages existing transformer-based Natural Language Inference (NLI) models and a frugal annotation strategy grounded in the well-established statistical heuristic of Probabilistic Bisection. Representing categories of interest as arbitrarily sophisticated taxonomies of statements then allows users to explicitly incorporate their own domain expertise into the classification system. The result can be described as a few-shot, active learning approach, which can easily be adapted for a range of use cases. All in all, this approach is designed to provide users with significant control over the output of the system, while keeping data annotation at a minimum. 

We benchmark this methodology against several baselines, including a fine-tuned large language model (LLM) and a zero-shot classification approach, in the context of three classification tasks using user-generated Web data: 1) fine-grained climate change contrarianism detection, 2) topic and stance detection, and 3) depression-related symptoms detection. Finally, we run experiments designed to analyse the output of the methodology, and to demonstrate how it can be adapted to the needs of researchers. We release this tool as part of a publicly available GitHub repository\footnote{\url{https://github.com/s-l-chausson/easyclaimsdetection}}. 

\section{Related work}\label{sec:background}

Since the advent of transformers \cite{vaswani2017attention}, text classification has been dominated by the ``pre-train/fine-tune" paradigm, where a pre-trained transformer-based Large Language Model (LLM) is further trained on a dataset annotated for the task and domain at hand. While the benefits of this paradigm have been significant, one limitation is the need for an annotated dataset, typically containing a few thousands examples, and the associated costs and labour to create it. 

In order to minimise the need for annotated data, different literatures have looked at leveraging the capabilities of large pre-trained language models in zero- or few-shot classification contexts, i.e. without any fine-tuning. One such approach is \emph{prompting} \cite{10.1145/3560815} and consists in re-framing a textual classification task into a \emph{Natural Language Generation} task. For instance, for topic classification, the prompt \emph{``This text is about..."} can be appended to some input text and fed to the language model. The word(s) generated by the model following the prompt then corresponds to the predicted topic. Prompting has been used for a range of applications, from aspect-based sentiment analysis \cite{li2021sentiprompt} to de-biasing \cite{schick2021self}. It has become particularly prevalent with the release of models such as GPT-4 by OpenAI \cite{achiam2023gpt}, Llama 2 by Meta \cite{touvron2023llama} or PaLM by Google \cite{chowdhery2023palm}. Unfortunately, only a few organisations, typically private companies, have the resources to train such models and often only provide access to the model via an API. In other words, using these models often requires transmitting one's data to these companies. Moreover, even models that \emph{are} made open-source are often difficult to run locally due to their size.

Another zero-shot classification approach, on which our own work builds, can be described as \emph{entailment-based}. It consists in re-framing a classification task as a \emph{Natural Language Inference} (NLI) task. NLI, in its simplest form, is a binary classification task: given a text $t$ and a hypothesis $h$ as input, the goal is to determine whether or not $h$ is entailed or supported by $t$ \cite{zeng2021automated}. Within this paradigm, for instance, each of the following three hypotheses can be appended to the input text one at a time and fed to an NLI model: \textit{``This text is about politics"}, \textit{``This text is about sports"} and \textit{``This text is about health"} \cite{yin-etal-2019-benchmarking}. The text is then classified as relating to a topic if the NLI model returns an entailment score greater than a given threshold (e.g. 0.5) for the corresponding hypothesis. 

When used in a zero-shot fashion, an NLI model's performance can drop as a result of the domain shift that occurs when applying it to data that significantly differs from the original training data. Moreover, modern neural networks have been shown to be miscalibrated \cite{guo2017calibration}: i.e. a neural network's confidence in its prediction generally does not reflect the \emph{likelihood} of this prediction being true. These two known limitations -- domain shift and model miscalibration -- indicate that applying a single threshold of 0.5 on the scores returned by an NLI model might not be a reliable heuristic in the context of zero-shot classification. 

To address the domain-shift problem, few-shot variants of entailment-based zero-shot classification approaches have been proposed. \citet{yin-etal-2020-universal}, for instance, augment existing NLI datasets with a few domain-specific examples of entailment, and fine-tune a somewhat ``specialised" NLI model on this new dataset. As opposed to this approach, our own methodology does not require fine-tuning a new NLI model. Instead, the small sample of data we annotate is used to train a separate parameter, i.e. a threshold, which is applied to NLI scores as a post-processing step. This makes our approach computationally less intensive, and arguably easier to grasp conceptually for researchers outside the field of Computer Science.

To address the general problem of miscalibration, on the other hand, a common solution is to transform the original predicted scores into calibrated probabilities \emph{post-hoc}. This is usually achieved by training a simple model, or ``scaler", on annotated data that differs from the training set. One such method, which has been found to be particularly effective for calibrating the predictions of neural networks, is \emph{Temperature Scaling} \cite{guo2017calibration}. Our own methodology can be seen as related to the literature on re-calibration in the sense that it uses a threshold-tuning step, whose output is applied \emph{post-hoc}. Its main added value, however, is that it does not rely on the availability of annotated data for the calibration. Rather, the threshold-tuning strategy we propose offers a pro-active solution to sampling data for annotation. Moreover, this strategy can be described as rooted in Active Learning principles \cite{ren2021survey} in the sense that it is explicitly designed to minimise the overall number of annotations. 

The latter is achieved by adapting an algorithm initially designed to solve one-dimensional root-finding problems, called the Probabilistic Bisection Algorithm (PBA) \cite{horstein1963sequential, frazier2019probabilistic, burnashev1974interval}. The PBA and its variants have been used for a range of applications in theoretical computer science \cite{pelc1989searching}, engineering \cite{castro2008active}, computer vision \cite{golubev2003sequential}, system simulation \cite{rodriguez2015information} and information theory \cite{tsiligkaridis2014collaborative}. As far as we are aware, however, our work is the first to use Probabilistic Bisection in the context of textual classification. 

Generally speaking, our methodology's reliance on taxonomies of claims is very much in line with existing work on \emph{theory-grounded} classification. One particularly relevant example from this literature is the approach proposed by \citet{sen2022depression}. In this paper, the authors use the Sentence-BERT \cite{reimers-gurevych-2019-sentence} model to embed online reviews from the Glassdoor platform, as well as statements such as ``The stress of my job caused me to have sleep problems" taken from the Occupational Depression Inventory \cite{bianchi2020occupational}, a questionnaire used by mental health professionals for the diagnostic of occupational depression. They then compute the cosine similarity between each (embedded) statement from the ODI and (embedded) online review, and calculate a threshold for each statement by annotating a stratified sample of examples. These thresholds are then used to turn cosine similarity scores into a binary classification. While very similar to this approach, our own method improve on it by 1) replacing cosine similarity by entailment scores and 2) proposing an Active-Learning strategy for determining thresholds.

\section{Method}\label{sec:method}

We divide our proposed methodology (see Figure \ref{fig:teaser}) into four steps: 1) \textbf{Taxonomy Definition}, 2) \textbf{Natural Language Inference}, 3) \textbf{Threshold-tuning} and 4) \textbf{Classification}. In what follows, we describe each of these steps in greater detail. 

\subsection{Taxonomy Definition}\label{subsec:claims_comp}

The first step of our proposed methodology consists in formulating a list of claims (e.g. ``Sea levels are not rising", ``Feminism is important", ``I feel sad"), to represent relevant classes. Each of these claims could be its own class. However, more elaborate logical relations between claims and classes can be defined. For instance, we can decide that a text should be classified as a particular class if it contains \emph{at least} one claim from a list. In other cases, several claims might need to appear \emph{together} in the same text to denote the class. Finally, we can also define the presence of a claim to signify the \emph{absence} of a class. More generally, arranging relevant claims into such taxonomies allows the user to define arbitrarily complex and fine-grained classes from claims, which provides them with a lot of flexibility and control over the scope of the classification. The claims themselves can be obtained from an existing theoretical framework (e.g. the BDI-II inventory for depression detection) or via a more inductive process of reviewing and synthesising relevant examples from the corpus of interest. 

\subsection{Natural Language Inference}\label{subsec:nli}

Having defined the list of claims that represent our classes, and their relation to each other, the second step of the methodology is to detect these claims in our data using an NLI model. The NLI model we use in this paper is BART\textsubscript{MNLI}, a version of the BART transformer-based model \cite{lewis-etal-2020-bart} fine-tuned on the Multi-genre Natural Language Inference (MNLI) dataset \cite{N18-1101}. This model, which is available via the HuggingFace platform, returns an entailment score that lies between 0 and 1 for the text/claim pair given as input. 

\subsection{Threshold-tuning}\label{subsec:thresholds}

Each combination of input text $d_i$ from our corpus and claim $c_j$ from our taxonomy now has an associated entailment score between 0 and 1. We assume that a higher entailment score means that $d_i$ is more likely to entail $c_j$. If the entailment score is over some threshold the text should be classified as entailing the claim. We want this threshold to be optimal, in the sense that it maximises the number of correct entailment classifications and minimises the number of incorrect ones. In practice, a unique threshold of 0.5 does not work well. Instead, a different threshold for each claim $c_j$ is required. We find each of these thresholds using an approach based on Bayesian inference, which involves asking a human to annotate the entailment for dynamically selected datapoints.

Formally, the threshold-finding task can be characterised as follows. Let the variable $s$ represent the entailment score for some input text $d_i$ and the claim $c_j$. Let $g$ then be the function that takes some value of $s$ as input. $g$ can be formalised as a monotonically increasing continuous function: $g : [0,1] \rightarrow \mathbb{R}$, such that $g$ crosses the $x$-axis at a single point $S^*$. The sign of $g$ gives us the class: if $s < S^*$ then $g(s) < 0$ and $c_j$ is \emph{not entailed} by the text, while if $s > S^*$ then $g(s) > 0$ and $c_j$ is \emph{entailed} by the text. In other words, $S^*$ is the optimal threshold for $c_j$. Our objective is to find $S^*$ but we do not know the exact function $g(s)$ and therefore cannot solve for $S^*$ analytically. We can however ask a human annotator to tell us whether a particular text $d_i$ with an entailment score equal to $s_t$ does indeed entail $c_j$, or not. This gives us the sign of $g$ at $s_t$, which tells us whether $S^*$ is greater or smaller than $s_t$. Finally, we assume that the sign of $g(s_t)$ given to us at each annotation is only correct with probability $p$, and incorrect with probability $q=p-1$, where $p$ is a hyper-parameter. This uncertainty is intended to capture both the annotator's subjectivity and the noise associated to the threshold $S^*$: after all, while $S^*$ should maximise correct entailment classifications and minimise incorrect ones, it is unlikely to eliminate errors altogether. 

\begin{figure*}[]
 \centering
 \makebox[\linewidth][c]{%
 \begin{subfigure}{0.25\textwidth}
  \centering
  \includegraphics[width=\linewidth]{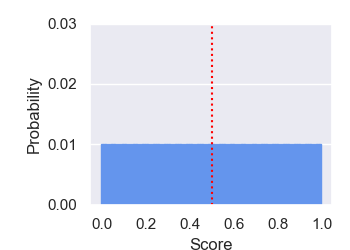}
  \caption{$1^{st}$ annotation}
  \label{fig:round_1}
 \end{subfigure}
 \begin{subfigure}{0.25\textwidth}
  \centering
  \includegraphics[width=\linewidth]{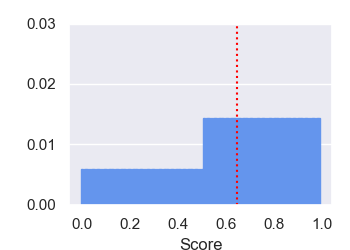}
  \caption{$2^{nd}$ annotation}
  \label{fig:round_2}
 \end{subfigure}
 \begin{subfigure}{0.25\textwidth}
  \centering
  \includegraphics[width=\linewidth]{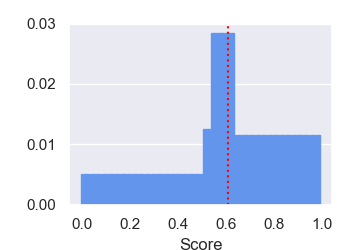}
  \caption{$4^{th}$ annotation}
  \label{fig:round_2}
 \end{subfigure}
 \begin{subfigure}{0.25\textwidth}
  \centering
  \includegraphics[width=\linewidth]{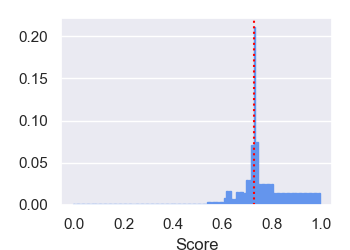}
  \caption{$16^{th}$ annotation}
  \label{fig:round_3}
 \end{subfigure}
 }
\caption{Probability distribution over threshold location for the claim \emph{1.1.2.0: ``Greenland is gaining ice"}. The red line shows the median of the current distribution, which defines the next datapoint to be selected for annotation. Over successive iterations of the Probabilistic Bisection Algorithm, the probability mass concentrates around the optimal threshold for that claim. Note that the y-axis for the $16^{th}$ annotation is different from other figures.} 
\label{fig:bisection_evol}
\end{figure*}

This formalisation frames our threshold-finding task as the well-studied mathematical problem of probabilistic root-finding, which can be solved with the least number of annotations using the Probabilistic Bisection Algorithm \cite{horstein1963sequential, frazier2019probabilistic}. This algorithm, once adapted to our needs, works as follows: 
\begin{enumerate}[itemsep=0.2pt,topsep=2pt]
    \item At timestep $t=0$, the location of our optimal threshold $S^*$ can be represented using a discretised uniform probability distribution $f_0$ ranging from 0 to 1 (Figure \ref{fig:round_1}). 
    \item At each subsequent timestep $t$, we find the median $m_t$ of $f_t$. 
    \item We then sample the text $d_t$ with entailment score $s_t$ closest to $m_t$ and find the sign of $g(s_t)$ by asking the human annotator whether $d_t$ entails $c_j$. 
    \item If $d_t$ does not entail $c_j$, then we know $S^* \geq s_t$ with probability $p$ and $S^* < s_t$ with probability $q$. We therefore update $f_t$ according to the following equation (Figure \ref{fig:round_2}), where $F_t$ stands for the cumulative distribution function of $f_t$:
    \begin{equation}
        f_{t+1} =
        \begin{cases}
            \frac{p}{F_t(s)}f_{t}(s), & \text{for } s \geq s_t\\
            \frac{q}{F_t(s)}f_t(s), & \text{for } s < s_t
        \end{cases}
        \label{eq:update_1}
    \end{equation}
    Inversely, if $d_t$ entails $c_j$, we know that $S^* < s_t$ with probability $p$ and  $S^* > s_t$ with probability $q$:
    \begin{equation}
            f_{t+1} = 
            \begin{cases}
                \frac{q}{F_t(s)}f_t(s), & \text{for } s \geq s_t\\
                \frac{p}{F_t(s)}f_t(s), & \text{for } s < s_t
            \end{cases}
            \label{eq:update_2}
    \end{equation}
\end{enumerate}

Operations 2, 3 and 4 can be repeated $n$ times until the probability interval for $S^*$ is considered sufficiently narrow by the user of the system (e.g. when 95\% of the probability mass is contained within a range of 0.10), or once no datapoint $s_t$ close enough to the median $m_t$ can be found due to data sparsity\footnote{Equations \ref{eq:update_1} and \ref{eq:update_2} set a natural limit on how much $s_t$ and $m_t$ can diverge. This is because $f_t$ needs to be scaled \emph{up} in the direction of the root, which means that $p$ must be greater than the probability mass $F_t(s)$ dividing it. As a result, if $s_t$ is so far off from $m_t$ that the probability mass of $f_t$ on the side of $s_t$ that needs to be scaled up is greater than $p$, then the algorithm stops.}. The optimal threshold is then set as the median of this final distribution $f_n$ (see Figure \ref{fig:round_3}).

We repeat this process for each claim from our taxonomy, such that each has an associated threshold.

\subsection{Classification}\label{subsec:classification}

The final step of the method consists in classifying each datapoint in the dataset into the correct class(es). To do this, we first label each datapoint with the claims it contains by applying the thresholds obtained during the \textbf{Threshold-tuning} step to the entailment scores obtained during the \textbf{Natural Language Inference} step. We then use these claim-labels and their relations to higher-level classes (defined during the \textbf{Taxonomy Definition} step) to assign the right class(es).

\section{Evaluation}

In this section, we illustrate and evaluate our methodology against four baselines on three different textual classification tasks for user-generated web-content, to demonstrate its versatility: 1) climate change contrarianism detection, 2) topic and stance classification, and 3) depressive symptoms detection. In what follows, we describe these tasks and datasets, the NLI model and PBA hyperparameters used to run our own methodology, and our choice of baselines. We then report and compare the performance of all approaches.

\subsection{Tasks, datasets and taxonomies}

\subsubsection{Fine-grained climate change contrarian claim detection}

The first task is taken from \citet{coan2021computer}. In this paper, the author provide a corpus of short newspaper and social media excerpts annotated with the climate change contrarian class (e.g. \emph{``Global warming is not happening"}, \emph{``Human greenhouse gases are not causing global warming"}) and sub-class they each pertain to (if any).

For simplicity, we limit our evaluation to only one of the five general classes from \citet{coan2021computer}: \emph{``Global warming is not happening"}. The sub-classes related to this general class are \emph{``1\_1: Ice/permafrost/snow cover isn’t melting"}, \emph{``1\_2: We are heading into an ice age"}, \emph{``1\_3: The weather is cold"}, \emph{``1\_4: There is a hiatus in warming"}, \emph{``1\_6: Sea level rise is exaggerated"} and \emph{``1\_7: Extreme weather is not increasing"}\footnote{The sub-class \emph{``1\_5: Oceans are cooling"} was included in the original taxonomy by \citet{coan2021computer} but not in their dataset, so we ignore it.}. In our taxonomy, each of these sub-classes corresponds to a claim. If the claim contains backslashes (``/"), we rephrase it into several independent claims. Finally, we paraphrase some of these claims to increase the number of claims in our taxonomy. An extract is then defined to pertain to the original sub-class if it contains any single one of its related claims. The resulting taxonomy of claims per sub-class is shown in Table \ref{tab:clim_taxonomy} of the Appendix. 

Defining the classes as such turns the initial multi-class classification problem (i.e. one class per datapoint) into a multi-label classification problem (i.e. one or more classes per datapoint). We believe this is a well-motivated change as paragraphs can contain claims pertaining to multiple classes. This was evident when we manually reviewed the dataset. For instance, the text:
\begin{quote}
    \textit{``JAXA shows ice increasing. At current rates of increase, the world will be covered with ice within 10 years."}
\end{quote}
...was initially only annotated with class \emph{1\_1: ``Ice/permafrost/snow cover isn't melting"}. However it also seems to fit the class \emph{1\_2: ``We're heading into an ice age"}. We therefore augmented the annotations for the entire test set\footnote{We did not re-annotate paragraphs originally annotated with the $0.0$ class as the switch from multi-class to multi-label should not impact them.} For the training set, we simply annotated datapoints selected by the PBA during the threshold-tuning step. The annotation augmentation was carried out by ourselves (the researchers), while taking into account the annotation instructions provided by \citet{coan2021computer}.

\subsubsection{Topic and stance classification}

The second task on which we evaluate our methodology is a topic and stance classification task published at SemEval 2016 \cite{mohammad-etal-2016-semeval}. Here, the aim is to classify which topic a tweet relates to (i.e. atheism, climate change, feminism, Hillary Clinton or abortion), and its stance with respect to this topic (``in favor", ``against" or ``neutral"). 

We carry out the topic and stance classification separately. Concretely, this means first using our methodology with a taxonomy of claims designed to determine which of the five topics each extract relates to: e.g. ``This text is about atheism", ``This text is about Hillary Clinton" (see Table \ref{tab:topic_taxonomy} in the Appendix for the full taxonomy). Unlike for the climate change contrarianism detection task, these claims are not paraphrases of the content of the tweet but rather meta-claims \emph{about} the tweets. In order to make the topic classification multi-class, we need to make scores for different claims comparable. To achieve this, we apply the following piece-wise linear transformation so that, for each claim, NLI scores lower than the claim's threshold are normalised to be contained in a range of 0 to 0.5, and score higher than the threshold are normalised to be between 0.5 and 1: 
\begin{equation}
        f(x, t_c) = 
        \begin{cases}
            \frac{0.5x}{t_c}, & \text{for } s \leq t_c\\
            0.5 + \frac{0.5(x - t_c)}{1 - t_c}, & \text{for } s > s_t
        \end{cases}
        \label{eq:linear_piecewise}
\end{equation}
Here, $x$ is the NLI score for an extract/claim input pair and $t_c$ is the threshold for that particular claim. The extract's topic is then defined as that for which the average (normalised) NLI score, calculated from all this topic's claims, is the greatest. 

Once the topic of an extract has been determined, our methodology is used again with a second taxonomy, designed to capture the stance of the tweet with respect to the topic. About half of the claims in this second taxonomy are related to an ``Against" stance (e.g. ``God exists", ``Atheism is bad" for the ``Atheism" topic), while the other half is related to an ``In favor" stance (e.g. ``God does not exist", ``Religion is a lie", ``I am an atheist", see Table \ref{tab:stance_taxonomy} in the Appendix for the full taxonomy). This taxonomy was obtained inductively by reading over a small number of extracts for each stance. If no ``Against" or ``In favor" claims are detected for the topic at hand, the extract is classified as having a ``Neutral" stance. If, on the other hand, more ``Against" claims are detected than ``In favor" claims, the stance is ``Against", and inversely. Finally, if the same (non-zero) number of ``Against" and ``In favor" claims are detected, Equation \ref{eq:linear_piecewise} is applied to the NLI score for all stance claims and the tweet is assigned the stance with the highest (normalised) average score. 

\subsubsection{Fine-grained depressive symptoms detection}

Finally, for the third task we use the BDI-Sen dataset put forward by \citet{perez2023bdi}. The objective here is to classify whether or not a user-generated sentence from the Reddit social media platform is indicative of the author suffering from one or more symptoms of depression, from a total of twenty-one possible symptoms . The symptoms, which include for instance ``Suicidal ideation", ``Irritability", or ``Indecisiveness", and are taken directly from the BDI-II, an inventory widely used for the diagnosis of depression by mental health professionals. 

The BDI-II itself consists of statements that capture each of these symptoms and for four different levels of severity, with 0 corresponding to the symptom not manifesting and 3 to it manifesting most strongly. For instance, the claims ``I feel the same about myself as ever", ``I have lost confidence in myself", ``I am disappointed in myself", ``I dislike myself" all capture the symptom ``Self-dislike" with increasing levels of severity. To build our taxonomy, we therefore simply take the claims corresponding to the severity levels 1, 2 and 3 for all symptoms (see the Appendix for the full taxonomy of claims). We then consider that a sentence is relevant to a symptom if it makes any of the claims related to that symptom.

\subsection{Running our methodology}

For this particular implementation of our methodology, we use BART\textsubscript{MNLI} as our NLI model. We use our adapted Probabilistic Bisection Algorithm (PBA) with a value of $p=0.7$ (we experiment with different values of $p$ in Experiment 1 later on). We set up the PBA so that the annotation process continues until no unannotated datapoint can be found close enough to the median due to data sparsity. We consider that the threshold-tuning for a given claim is ``complete" if 95\% of the probability mass is concentrated in a range narrower than 0.20 by the last annotation, otherwise it is labelled as an ``early stop". Table \ref{tab:threshld_tuning_stats} summarises the number of claims for each task, the average number of annotations per claim, the average width of the 95\% confidence interval around the threshold at the final annotation and the range of thresholds obtained. These results are reported for all claims, and for ``complete" and ``early stop" claims separately.

\begin{table}[]
\centering
\scriptsize
\begin{tabular}{llcccc}
\cline{3-6}
 &  & \textbf{\begin{tabular}[c]{@{}c@{}}Nb\\ claims\end{tabular}} & \textbf{\begin{tabular}[c]{@{}c@{}}Avg nb\\ annots\end{tabular}} & \textbf{\begin{tabular}[c]{@{}c@{}}Avg 95\%\\ CI width\end{tabular}} & \textbf{\begin{tabular}[c]{@{}c@{}}Threshold\\ range\end{tabular}} \\ \hline
\multirow{3}{*}{\textbf{\begin{tabular}[c]{@{}l@{}}Climate \\ change\end{tabular}}} & All & 31 & 19.4 & 0.37 & {[}0.34, 0.99{]} \\
 & Complete & 9 & 19.6 & 0.12 & {[}0.86, 0.99{]} \\
 & Early stop & 22 & 19.3 & 0.48 & {[}0.34, 0.93{]} \\ \hline
\multirow{3}{*}{\textbf{\begin{tabular}[c]{@{}l@{}}Topic/\\ stance\end{tabular}}} & All & 85 & 20.5 & 0.41 & {[}0.03, 0.99{]} \\
 & Complete & 15 & 26.3 & 0.11 & {[}0.03, 0.99{]} \\
 & Early stop & 70 & 19.3 & 0.47 & {[}0.10, 0.95{]} \\ \hline
\multirow{3}{*}{\textbf{\begin{tabular}[c]{@{}l@{}}Depres-\\sion\end{tabular}}} & All & 64 & 12.1 & 0.21 & {[}0.52, 0.99{]} \\
 & Complete & 46 & 14.1 & 0.08 & {[}0.94, 0.99{]} \\
 & Early stop & 18 & 7.1 & 0.55 & {[}0.52, 0.95{]} \\ \hline
\end{tabular}
\caption{Number of claims, average number of annotations per claim, average width of the 95\% confidence interval around the threshold at the final annotation and range of thresholds obtained when using our methodology for the classification tasks.}
\label{tab:threshld_tuning_stats}
\end{table}

\begin{table*}[h!]
\centering
\scriptsize
\begin{tabular}{lcccccc}
\cline{2-7}
   & \multicolumn{2}{c}{\textbf{\begin{tabular}[c]{@{}c@{}}Climate change\end{tabular}}} & \multicolumn{2}{c}{\textbf{\begin{tabular}[c]{@{}c@{}}Topic/ Stance\end{tabular}}} & \multicolumn{2}{c}{\textbf{\begin{tabular}[c]{@{}c@{}}Depression\end{tabular}}} \\ \hline
\multicolumn{1}{c}{\textbf{Approach}} & \textbf{F1-score}   & \textbf{\# Datapoints} & \textbf{F1-score}  & \textbf{\# Datapoints}  & \textbf{F1-score} & \textbf{\# Datapoints}                              \\ \hline
Fine-tuned BERT  & 0.62 & 23,436 & \textbf{0.55} & 2,525 & 0.30 & 1,060 \\
Cosine similarity w. threshold-tuning & 0.47 & 680 & 0.49 & 1,271 & 0.21 & 241 \\ 
Entailment as Zero-Shot Classification  & 0.09 & 0 & 0.47 & 0 & 0.07 & 0 \\
Entailment w. Temperature Scaling & 0.58 ($N=20$) & 298 & 0.44 ($N=10$) & 2,525 & 0.23 ($N=160$)  & 1,006 \\
Prompting w. Llama 2 & 0.11 & 2 & 0.08 & 4 & 0.28 & 2 \\ \hline
Entailment w. threshold-tuning (ours)  & \textbf{0.68}  & 457 & 0.54  & 1,112 & \textbf{0.35} & 277 \\ \hline
\end{tabular}
\caption{Weighted-average F1-score of our own methodology using the BART\textsubscript{MNLI} model compared to five baselines: 1) a BERT model fine-tuned on the entire training set, 2) a version of the methodology that uses the cosine similarity instead of an entailment scores, 3) a Zero-Shot Learning (ZSL) version of the methodology with a unique threshold of 0.5 for all claims, 4) a version of the methodology that uses Temperature Scaling instead of the threshold-tuning step and 5) a prompting approach that uses the Llama 2 model with 70 billion parameters. The table also provides the number of annotated datapoints required to obtain these F1-scores. The Llama 2 model could not easily be constrained into multi-class, rather than multi-label, classification for the Topic detection task. We therefore considered the topic to be correctly predicted by this baseline even if it also detected other topics.}
\label{tab:nli_models_comp_perf}
\end{table*}

\subsection{Baselines}\label{subsec:baselines}

We run five different baselines to validate different aspects of our methodology. To compare the method's capabilities with a ``traditional" supervised-learning approach, we fine-tune a version of the BERT model on the entire training set for each task. The model is fine-tuned using an early stopping paradigm: i.e. loss is measured at the end of every epoch on the validation set, and as long as this loss decreases, the model is fine-tuned for an other epoch. The model used then is that with the lowest validation loss\footnote{In the case of the depression dataset, we removed 203 datapoints from the training set which were also in the test set. Moreover, we up-sampled the training set such that every symptom label appears at least 50 times.}.

Secondly, to validate the value of using NLI as the underlying framework, we also compare our methodology to a version that uses cosine similarity instead of entailment scores. To do so, we use the Sentence-BERT model (SBERT) \cite{reimers-gurevych-2019-sentence} to obtain vector representations of our texts and claims, and measure the cosine similarity between each text/claim pair. We then use the PBA to find the optimal cosine similarity threshold for each claim, starting with a uniform probability distribution ranging from -1 to 1 instead of 0 to 1. This baseline, particularly when used for the Depressive Symptoms detection task with claims taken from the BDI-II inventory, is very similar to the approach proposed by \citet{sen2022depression}.

Thirdly, to validate the need to tune the thresholds for each claim, we run a version of our methodology where a unique threshold of 0.5 is used for all the claims. Because this baseline requires no annotations it can be describes it as ``zero-shot". 

To further validate the value of threshold-tuning, we compare our methodology with a fourth approach that also uses entailment but which, instead of the Probabilistic Bisection algorithm, uses Temperature Scaling. Concretely, for each claim, we sample $N$ data-points from the training set that contain the claim and $N$ that do not, and use this sample to train a Temperature Scaling calibrator. This is then used to re-calibrate this claim's NLI scores. We repeated this process for $N \in \{5, 10, 20, 40, 80, 160\}$. In our results, we only report the F1-score obtained with the smallest value of $N$ that leads to the greatest performance.

Finally, we compare our methodology to a prompting approach using Llama 2 model with 7 billion parameters. For each inference, the prompt contains two examples from the training set with their corresponding label followed by the text to classify (see Table \ref{tab:prompt_examples} in the Appendix for the prompt template for each task).

\subsection{Results}

The test set performance per task of all baselines as well as our proposed methodology, with the total number of annotations they each required, is given in Table \ref{tab:nli_models_comp_perf}. We see that our proposed methodology using the BART\textsubscript{MNLI} model and threshold-tuning out-performs every baseline presented in the \textbf{Baselines} section for the Climate Change Contrarianism and Depressive Symptoms detection tasks. Moreover, while it does not out-perform the fine-tuned BERT baseline for the Topic/Stance classification task, it comes as a close second\footnote{While the datasets we used are annotated per class (e.g. ``Atheism, Against" or ``Sadness") they are not annotated for each of the claims we used individually. For simplicity, we assumed that all claims pertaining to a single class share the same annotation. This simplification might contribute to our method scoring lower than if each claim had been annotated independently.}. An equally important result to consider when comparing to the fine-tuned BERT baseline though is that our proposed methodology requires a lot less data to perform this well: 2.3 times less for topic/stance classification, 3.8 times less for the depressive symptoms detection and a dramatic 51.3 times less for the climate change contrarian claims detection. This last result ought to be taken with a pinch of salt however, as 21,628 datapoints out of the 23,436 in the training set correspond to paragraphs that make no climate change contrarian claim. Nevertheless, even if counting only the remaining (arguably more valuable) 1,808 datapoints, our methodology requires 4 times less data.

Comparisons with other baselines than the fine-tuned BERT also validate several aspects of our methodology. The use of an NLI model instead of cosine similarity between SBERT embeddings does contribute to a significantly improved performance. So does the shift from a zero-shot to a few-shot approach (e.g. using \emph{post-hoc} Temperature Scaling or threshold-tuning). Moreover, the low performance of the Zero-Shot baseline paired with the wide range of thresholds obtained with our method (see Table \ref{tab:threshld_tuning_stats}) illustrates how ill-suited assuming a ``one-size-fits-all" threshold of 0.5 is when using NLI models for textual classification. Finally, our own threshold-tuning approach leads to significantly better performance with generally fewer annotations than Temperature Scaling.

\section{Experiments}\label{sec:experiments}

In this section, we first evaluate the impact of the hyper-parameter $p$, used for Probabilistic Bisection during the threshold-tuning step, on final results (Experiment 1). We then evaluate to what extent the threshold-tuning algorithm converges to similar thresholds when used on separate splits of the same data (Experiment 2). Finally, we experiment with two strategies designed to improve the precision (Experiments 3) and recall (Experiment 4) respectively. For simplicity, we focus exclusively on the Climate Change Contrarianism dataset for these experiments.

\subsection{Experiment 1: Changing the value of $p$ during the Probabilistic Bisection}\label{subsec:exp_5}

The Probabilistic Bisection Algorithm we use to tune the threshold of each claim requires a hyper-parameter $p$, which captures the probability of the threshold being in the direction indicated by the annotator. In our initial set up, we set $p$ equal to 0.7. We hypothesise that the value of $p$ does not impact the final value of the threshold, but rather acts as a learning rate: the higher the value of $p$, the quicker the algorithm should converge on the threshold. 

\begin{figure*}[]
 \centering
 \makebox[\linewidth][c]{%
 \begin{subfigure}{0.25\textwidth}
  \centering
  \includegraphics[width=\linewidth]{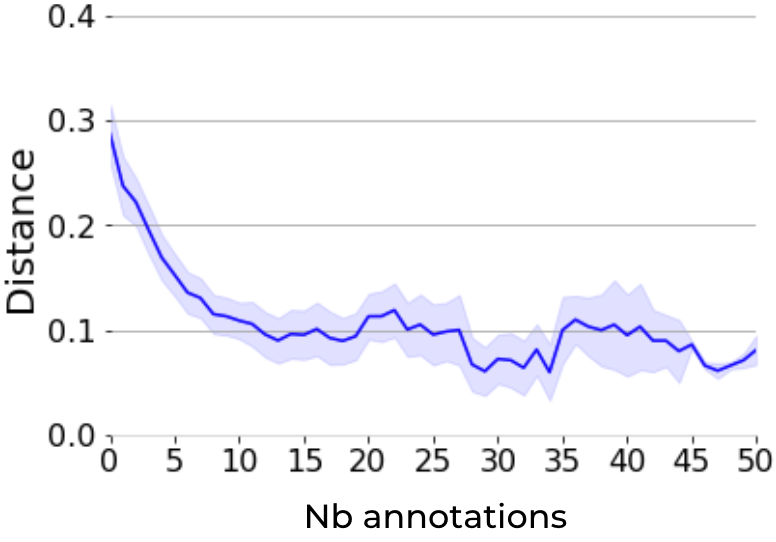}
  \caption{$p=0.6$}
  \label{fig:p_0.6}
 \end{subfigure}\hspace{2pt}
 \begin{subfigure}{0.23\textwidth}
  \centering
  \includegraphics[width=\linewidth]{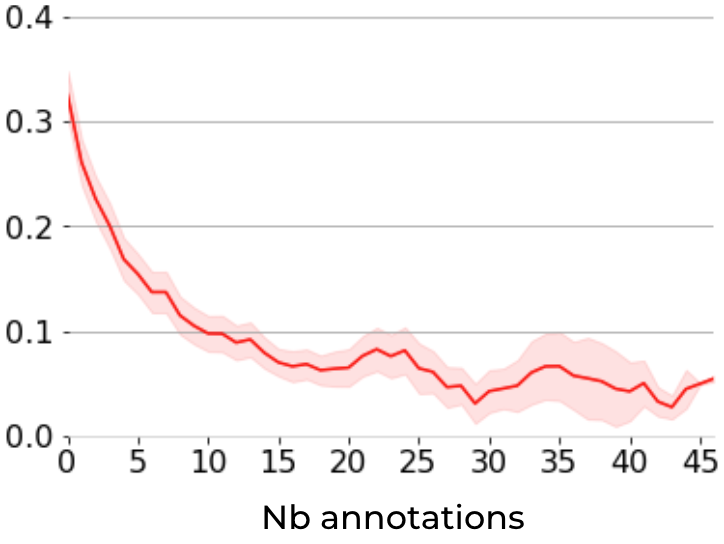}
  \caption{$p=0.7$}
  \label{fig:p_0.7}
 \end{subfigure}\hspace{2pt}
 \begin{subfigure}{0.235\textwidth}
  \centering
  \includegraphics[width=\linewidth]{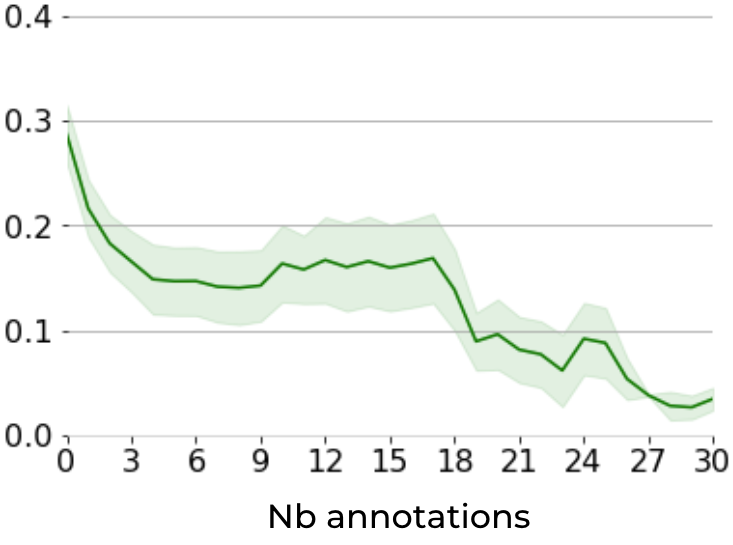}
  \caption{$p=0.8$}
  \label{fig:p_0.8}
 \end{subfigure}\hspace{2pt}
 \begin{subfigure}{0.23\textwidth}
  \centering
  \includegraphics[width=\linewidth]{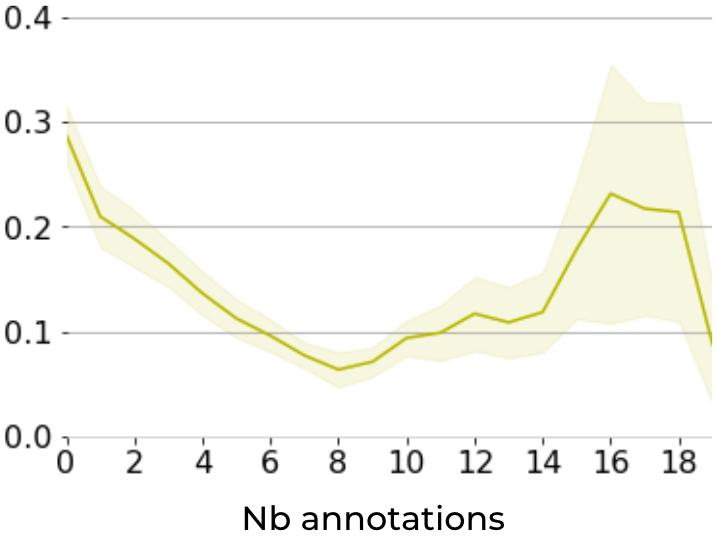}
  \caption{$p=0.9$}
  \label{fig:p_0.9}
 \end{subfigure}
 }
\caption{Average distance from the true optimal threshold, as calculated from the entire training set, per round of annotation and associated standard error. The standard error is calculated from the standard deviation from the average distance across all claims, and the number of claims still being annotated at that given round of annotation.} 
\label{fig:distance_from_threshold}
\end{figure*}

To test this hypothesis, we first calculate the threshold for each claim $c_j$ that maximises accuracy, using all the annotated datapoints from the training set. This is our ground truth optimal threshold, which the PBA is designed to approximate. We then perform the threshold-tuning step with $p$ equal to $0.6$, $0.7$ (our initial set up), $0.8$ and $0.9$. For each timestep $t$ of each annotation sequence, we calculate the distance of the estimated threshold from the ground truth threshold (i.e. the absolute value of their difference). For each value of $p$, we then plot the average distance per timestep, with the associated standard error (see Figure \ref{fig:distance_from_threshold}).

The resulting plots do suggest that the PBA does approach the ground truth threshold whether $p$ is equal to 0.6, 0.7 or 0.8. In each case, the algorithm reaches the average minimal distance from the threshold around the $30^{th}$ annotation, and then either plateaus (when $p=0.6$ and $0.7$), or stops altogether due to data sparsity (when $p=0.8$). This minimal distance is lesser for $p=0.7$ and $p=0.8$ at around $3$ percentage points, instead of $6$ for $p=0.6$. 

When $p=0.9$, however, the threshold diverges away from the optimal threshold rather erratically from the $8^{th}$ annotation. One possible interpretation of this trend is that the greater the value of $p$, the more weight will be given to the latest annotation compared to previous annotations, whose combined effect is captured by the current probability distribution. As $p$ becomes too great, however, this effect becomes disproportionate: the latest annotation essentially single-handedly shifts most of the probability mass in the direction it favours. This makes the algorithm very sensitive to noise, which is particularly likely to occur close to the threshold. 

Together, these observations provide only partial support for our hypothesis. While the exact value of $p$ might not be critical, very high and low values of $p$ should be avoided. A value of $p$ from the $0.7$ to $0.8$ range appears to be a good default. More generally, the assumption that $p$ should remain constant throughout the annotation of a claim is a rather strong simplification, and an aspect of our threshold-tuning method to potentially improve on. A better approach, which we leave for future work, would re-evaluate the value of $p$ dynamically at each timestep, for instance based on how neatly the current median of the probability distribution separates positive and negative examples from previous annotations, and how confident the annotator feels about each annotation. 

\subsection{Experiment 2: Threshold-tuning on separate folds of the data}\label{subsec:exp_2}

When running the threshold-tuning, several datapoints with equal scores can theoretically be selected for annotation. While the annotations for these different datapoints might individually differ, the PBA should in theory converge to similar thresholds. In this experiment, we attempt to evaluate to what extent this is the case. 

To do so, we split the Climate Change Contrarianism dataset into three folds of equal size (7,812 datapoints per fold) and run the threshold-tuning independently on each fold. Table \ref{tab:distance_from_threshold} reports the average difference between the highest and lowest threshold obtained for all claims, and the standard deviation of this difference. The ``Complete", ``Early stop" and ``Mixed" categories in the table correspond to claims for which the threshold-tuning ran to completion on \emph{all}, \emph{none} and \emph{some but not all} of the three folds. 

As we can see, the difference in the thresholds obtained ranges from 0.01 to 0.28. However, this difference is much smaller when the threshold-tuning for a claim has run to completion for all three folds (at most 0.12) than when it has run to completion for only some or none of the folds. This result highlights the importance of letting the algorithm run to ``completion" to obtain more trustworthy thresholds, i.e. thresholds that are similar even when obtained by annotating different individual datapoints. It also highlights that looking at the width of the confidence interval generated by the PBA, as we have, is indeed an informative heuristic for estimating the trustworthiness of the threshold.

\begin{table}[h!]
\centering
\begin{tabular}{lllll}
\cline{2-5}
 & \textbf{Min} & \textbf{Max} & \textbf{Avg} & \textbf{Std} \\ \hline
\textbf{All} & 0.01 & 0.28 & 0.13 & 0.07 \\
\textbf{Complete} & 0.01 & 0.12 & 0.07 & 0.04 \\
\textbf{Early stop} & 0.05 & 0.28 & 0.16 & 0.06 \\
\textbf{Mixed} & 0.02 & 0.11 & 0.07 & 0.03 \\ \hline
\end{tabular}
\caption{Difference between the highest and lowest threshold obtained for the same claim from three separate folds of the Climate Change Contrarianism dataset.} 
\label{tab:distance_from_threshold}
\end{table}

\subsection{Experiment 3: Improving precision using negated claims}\label{subsec:exp_3}

We hypothesise that one possible reason our methodology labels datapoints with claims they do not contain (i.e. false positives) is due to a poor understanding of negations from the BART\textsubscript{MNLI} model. For instance, it is possible that paragraphs containing the claim \emph{``Glaciers are vanishing"} or \emph{``Antarctica is not gaining ice"} are incorrectly labelled as containing the claims \emph{``Glaciers are not vanishing"} and \emph{``Antarctica is gaining ice"} respectively, due to the similarity of these statements apart from the negation. 

To test this hypothesis, we write a negated version $\neg c_j$of each claim $c_j$, i.e. a version that expresses a climate change believing stance rather than a climate change contrarian stance (see Table \ref{tab:clim_taxonomy} in the Appendix for the full list of negated claims), and run the BART\textsubscript{MNLI} model for these negated claims on our test set. For a piece of text $t$ to contain the claim $c_j$, we then require the entailment score between $t$ and $c_j$ to be greater than the entailment score between $t$ and $\neg c_j$. If the hypothesis is correct, this should improve precision on the test set. The results of this experiment, with a breakdown per class, are shown in Table \ref{tab:exp_2a}.

\begin{table}[h!]
\small
\centering
\begin{tabular}{cccc}
\hline
\textbf{Class}     & \textbf{Precision}                   & \textbf{Recall}                      & \textbf{F1-score}                    \\ \hline
1.1                & \cellcolor[HTML]{FBFAFF}0.70 (+0.02) & \cellcolor[HTML]{FFF0F0}0.58 (-0.06) & \cellcolor[HTML]{FEFAF9}0.64 (-0.02) \\
1.2                & \cellcolor[HTML]{FFFFFF}0.82 (+0.00) & \cellcolor[HTML]{FFFFFF}0.75 (+0.00) & \cellcolor[HTML]{FFFFFF}0.78 (+0.00) \\
1.3                & \cellcolor[HTML]{FFFCFB}0.60 (-0.01) & \cellcolor[HTML]{FEFAF9}0.75 (-0.02) & \cellcolor[HTML]{FFFCFB}0.67 (-0.01) \\
1.4                & \cellcolor[HTML]{FBFAFF}0.82 (+0.02) & \cellcolor[HTML]{FFFFFF}0.64 (+0.00) & \cellcolor[HTML]{FCFBFE}0.72 (+0.01) \\
1.6                & \cellcolor[HTML]{F1F0FF}0.94 (+0.06)  & \cellcolor[HTML]{FFFFFF}0.62 (+0.00) & \cellcolor[HTML]{FBFAFF}0.75 (+0.02) \\
1.7                & \cellcolor[HTML]{FCFBFE}0.52 (+0.01) & \cellcolor[HTML]{FFFFFF}0.65 (+0.00) & \cellcolor[HTML]{FCFBFE}0.58 (+0.01) \\ \hline
\textbf{Weighted avg} & \cellcolor[HTML]{FCFBFE}0.73 (+0.02) & \cellcolor[HTML]{FFFCFB}0.66 (-0.01) & \cellcolor[HTML]{FFFFFF}0.68 (+0.00) \\ \hline
\end{tabular}
\caption{Fine-grained test set performance of our proposed methodology when only considering claims with a higher entailment score than their negated counterparts (with comparison to the initial performance in percentage points).}
\label{tab:exp_2a}
\end{table}

Imposing this extra constraint does marginally improve precision for several classes, namely 1.1, 1.4 and 1.6. For class 1.1, however, this improvement is accompanied by a decrease in recall. Moreover, in the case of class 1.3, using the negated claims slightly decreases both precision and recall. The latter might be due to the fact that some of the claims we formulated for this class, and their negated counterparts, are particularly long and complex. For instance, claim 1\_3\_0\_0 \emph{``The weather is cold therefore global warming is not happening"} and its negated version \emph{``The weather being cold is not proof that global warming is not happening"} contain several verbs, subordinate clauses (\emph{``...that global warming is not happening"}), conditional relations (\emph{``The weather is cold \textbf{therefore}..."} and double negations (\emph{``is \textbf{not} proof that global warming is \textbf{not} happening"}). It is possible that this complexity confounds the BART\textsubscript{MNLI} model, suggesting that simpler formulations could improve the results.

\subsection{Experiment 4: Improving recall by adding claims}\label{subsec:exp_4}

Having investigated whether using negated claims can improve precision, we will now investigate whether recall can be increased by including more claims to describe each class. After manually reviewing climate change contrarian extracts from the training set undetected by our method given our initial list of claims, we add 10 new claims to our list. These either capture new ideas altogether (e.g. \emph{``Sea level rise is offset by other factors"}), or are paraphrases of existing claims (see Table \ref{tab:clim_taxonomy} in the Appendix for the full list). We then ran the BART\textsubscript{MNLI} model on these new claims and obtain their respective thresholds using the same Probabilistic Bisection approach as before. The results of this experiment are shown in Table \ref{tab:exp_4}.

\begin{table}[h!]
\small
\centering
\begin{tabular}{cccc}
\hline
\textbf{Class}                         & \textbf{Precision}                   & \textbf{Recall}                      & \textbf{F1-score}                    \\ \hline

1.1 & \cellcolor[HTML]{fffafa}0.66 (-0.02) & \cellcolor[HTML]{f2f2ff}0.69 (+0.05) & \cellcolor[HTML]{fafaff}0.68 (+0.02) \\
1.2 & \cellcolor[HTML]{ffa8a8}0.48 (-0.34) & \cellcolor[HTML]{d4d4ff}0.92 (+0.17) & \cellcolor[HTML]{ffd8d8}0.63 (-0.15) \\
1.3 & \cellcolor[HTML]{fcfcff}0.62 (+0.01) & \cellcolor[HTML]{eaeaff}0.85 (+0.08) & \cellcolor[HTML]{f4f4ff}0.72 (+0.04) \\
1.4 & \cellcolor[HTML]{ffcccc}0.60 (-0.20) & \cellcolor[HTML]{d8d8ff}0.79 (+0.15) & \cellcolor[HTML]{fff8f8}0.68 (-0.03) \\
1.6 & \cellcolor[HTML]{ffe2e2}0.77 (-0.11) & \cellcolor[HTML]{e8e8ff}0.71 (+0.09) & \cellcolor[HTML]{fcfcff}0.74 (+0.01) \\
1.7 & \cellcolor[HTML]{ffdcdc}0.37 (-0.14) & \cellcolor[HTML]{f8f8ff}0.68 (+0.03) & \cellcolor[HTML]{ffe6e6}0.47 (-0.10) \\ \hline
\multicolumn{1}{l}{\textbf{Weighted avg}} & \cellcolor[HTML]{ffe8e8}0.59 (-0.12) & \cellcolor[HTML]{e6e6ff}0.77 (+0.10) & \cellcolor[HTML]{fff8f8}0.66 (-0.02) \\ \hline
\end{tabular}
\caption{Fine-grained test set performance of our proposed methodology when adding 10 extra claims (with comparison to the initial performance in percentage points)}
\label{tab:exp_4}
\end{table}

We see that, as expected, the impact of adding these new claims is an increase in recall for all classes. With the exception of classes 1.3, however, this increase in recall is accompanied by a decrease in precision. 
It is worth noting that, if one has access to a validation set, one can perform a ``grid search" to find the exact combination of claims that maximises a particular performance metric. A validation set might not be available for most use cases though, and therefore it is not required by our methodology.  

\section{Conclusion}\label{sec:conclusion}

In summary, we propose a qualitative and versatile methodology as a general paradigm for claim-based text classification. Two particularly important features of this methodology are 1) the focus on defining classes as taxonomies of claims, and 2) the use of the Probabilistic Bisection Algorithm as a few-shot, active learning strategy. We illustrate and evaluate the methodology in the context of climate change contrarianism detection, topic and stance classification, and depressive symptoms detection for user generated text from the Web. However we argue that it can generalise to any task requiring the matching of pre-defined claims with pieces of text. Researchers interested in deploying our methodology can use our publicly available GitHub repository\footnote{\url{https://github.com/s-l-chausson/easyclaimsdetection}}, which also provides code to replicate our experiments.

We observed in the context of our evaluation that, due to data sparsity, the PBA occasionally needs to stop before having satisfactorily converged on a threshold. Starting with a large pool of unannotated data can mitigate this, suggesting that our methodology is most appropriate when deployed on \emph{large} corpora. 
The problem of early-stopping is also mitigated by the PBA returning a final probability distribution, from which a confidence interval can be calculated. This confidence interval can then be used as a heuristic to estimate the ``trustworthiness" of a threshold. We believe more such quality heuristics could be developed. These would in turn further help users of our methodology identify highly and poorly performing claims respectively, iteratively refine their taxonomies, and ultimately improve the quality of the predicted labels \emph{without} a validation and/or test set. 
Another limitation of our methodology is that it assumes a constant value for the probability $p$ throughout the annotation process. A possible improvement would therefore be to account for dynamically evolving values of $p$. 
Finally, the BART\textsubscript{MNLI} model we use in this paper is already several years old and as such does not reflect the latest advances in training LLMs. Fine-tuning recent open-source LLMs (e.g. Llama 2) on the MNLI dataset, which as far as we are aware has not been done, and using it as part of our methodology could also improve performance.

Overall, we believe the method we propose, and the shift in paradigm that underlies it, can be beneficial for researchers in Computational Social Sciences and Web Content Analysis, but perhaps also more generally any research interested in the detection of claims in large corpora. First of all, the method drastically reduces the need for annotation compared to more traditional supervised-learning approaches. This makes it a much cheaper and quicker-to-deploy alternative. Secondly, while we realise that carefully hand-crafting classes from claims can seem somewhat against the tide of \emph{laissez-faire} end-to-end Deep Learning, where it is assumed that the model should be able to develop a latent understanding of the concepts most relevant to each class from the data directly (if provided with enough data), it is precisely this hand-crafting that makes our proposed methodology particularly actionable. If a piece of text is misclassified, researchers can trace back to claims that caused the error and act on the misclassification by modifying their taxonomy. This is valuable in the context of computational research, where rooting classes in a rigorous and expertly-informed theoretical framework is often essential to promoting trust in the technology. Finally, we believe that focusing on taxonomies of claims can facilitate the sharing of existing expertise across research groups and applications. After all, adapting a taxonomy to a new use case is arguably much easier than adapting a dataset. Moreover, writing new taxonomies can be done in collaboration with researchers with little to no experience in computer science or NLP, thus promoting valuable interdisciplinary work.

\section{Acknowledgements}

This work was supported in part by the UKRI Centre for Doctoral Training in Natural Language Processing, funded by the UKRI (grant EP/S022481/1) and the School of Informatics at the University of Edinburgh.

\bibliography{sample-base}

\appendix

\begin{table*}[]
\centering
\scriptsize
\begin{tabular}{ccll}
\hline
\textbf{Class} & \textbf{Claim ID} & \multicolumn{1}{c}{\textbf{Claim}} & \multicolumn{1}{c}{\textbf{Negated claim}} \\ \hline
1\_1 & 1\_1\_0\_0 & The world's ice is not melting & The world's ice is melting \\
 & 1\_1\_0\_1 & The permafrost is not melting & The permafrost is melting \\
 & 1\_1\_0\_2 & The world's snow cover is not melting & The world's snow cover is melting \\
\multicolumn{1}{l}{} & \multicolumn{1}{l}{1\_1\_0\_3*} & Ice thickness is not decreasing & N/A \\
\multicolumn{1}{l}{} & \multicolumn{1}{l}{1\_1\_0\_4*} & Snow cover is not decreasing & N/A \\
 & 1\_1\_1\_0 & Antarctica is gaining ice & Antarctica is losing ice \\
 & 1\_1\_1\_1 & Antarctica's ice is not melting & Antarctica's ice is melting \\
\multicolumn{1}{l}{} & \multicolumn{1}{l}{1\_1\_1\_2*} & Antarctica is cooling & N/A \\
 & 1\_1\_2\_0 & Greenland is gaining ice & Greenland is losing ice \\
 & 1\_1\_2\_1 & Greenland's ice is not melting & Greenland's ice is melting \\
 & 1\_1\_3\_0 & The Arctic sea ice is not vanishing & The Arctic sea ice is vanishing \\
 & 1\_1\_3\_1 & The Arctic sea ice is increasing & The Arctic sea ice is decreasing \\
\multicolumn{1}{l}{} & \multicolumn{1}{l}{1\_1\_3\_2*} & The Arctic is cooling & N/A \\
 & 1\_1\_4\_0 & Glaciers are not vanishing & Glaciers are vanishing \\
 & 1\_1\_4\_1 & Glaciers are increasing & Glaciers are decreasing \\ \hline
1\_2 & 1\_2\_0\_0 & We are heading into an ice age & We are not heading into an ice age \\
 & 1\_2\_0\_1 & We are heading into a period of global cooling & We are not heading into a period of global cooling \\
\multicolumn{1}{l}{} & \multicolumn{1}{l}{1\_2\_0\_2*} & Global temperatures are cooling & N/A \\ \hline
1\_3 & 1\_3\_0\_0 & The weather is cold therefore global warming is not happening & \begin{tabular}[c]{@{}l@{}}The weather being cold is not proof that global warming\\ is not happening\end{tabular} \\
 & 1\_3\_0\_1 & The weather is colder than in the past & The weather is not colder than in the past \\
 & 1\_3\_0\_2 & \begin{tabular}[c]{@{}l@{}}The weather is exceptionally cold therefore global warming is\\ not happening\end{tabular} & \begin{tabular}[c]{@{}l@{}}The weather being exceptionally cold is proof that global\\ warming is happening\end{tabular} \\
 & 1\_3\_0\_3 & It's snowing therefore global warming is not happening & \begin{tabular}[c]{@{}l@{}}The fact that it's snowing is not proof that global warming\\ is not happening\end{tabular} \\
 & 1\_3\_0\_4 & It's snowing more than in the past & It's not snowing more than in the past \\
 & 1\_3\_0\_5 & \begin{tabular}[c]{@{}l@{}}It's snowing an exceptional amount therefore global warming\\ is not happening\end{tabular} & \begin{tabular}[c]{@{}l@{}}The fact that it's snowing an exceptional amount is proof \\ that global warming is happening\end{tabular} \\
\multicolumn{1}{l}{} & \multicolumn{1}{l}{1\_3\_0\_6*} & It's snowing & N/A \\
\multicolumn{1}{l}{} & \multicolumn{1}{l}{1\_3\_0\_7*} & The weather is cold & N/A \\ \hline
1\_4 & 1\_4\_0\_0 & The climate has not warmed over the last few decades & The climate has warmed over the last few decades \\
 & 1\_4\_0\_1 & The climate has not changed over the last few decade & The climate has changed over the last few decade \\
\multicolumn{1}{l}{} & \multicolumn{1}{l}{1\_4\_0\_3*} & Global temperatures have not increased over the last few decades & N/A \\
\multicolumn{1}{l}{} & \multicolumn{1}{l}{1\_4\_0\_4*} & Data does not show a rise in global temperatures & N/A \\
\multicolumn{1}{l}{} & \multicolumn{1}{l}{1\_4\_0\_5*} & \begin{tabular}[c]{@{}l@{}}Temperature data from the last few decades contradicts global\\ warming predictions\end{tabular} & N/A \\
\multicolumn{1}{l}{} & \multicolumn{1}{l}{1\_4\_0\_6*} & There has been a hiatus in global warming for the last few decades & N/A \\ \hline
1\_6 & 1\_6\_0\_0 & The rise of sea level is exaggerated & The rise of sea level is not exaggerated \\
 & 1\_6\_0\_1 & The rise of sea level is not accelerating & The rise of sea level is accelerating \\
 & 1\_6\_0\_2 & Sea levels are not rising & Sea levels are rising \\
 & 1\_6\_0\_3 & The rise of sea level is normal & The rise of sea level is not normal \\
 & 1\_6\_0\_4 & Sea level rise is not related to climate change & Sea level rise is related to climate change \\
\multicolumn{1}{l}{} & \multicolumn{1}{l}{1\_6\_0\_5*} & Sea level rise is offset by other factors & N/A \\ \hline
1\_7 & 1\_7\_0\_0 & Extreme weather is not increasing & Extreme weather is increasing \\
 & 1\_7\_0\_1 & Extreme weather has happened before & Extreme weather has not happened before \\
 & 1\_7\_0\_2 & Extreme weather is not linked to climate change & Extreme weather is linked to climate change \\
 & 1\_7\_0\_3 & Extreme weather has always existed & Extreme weather has not always existed \\
\multicolumn{1}{l}{} & \multicolumn{1}{l}{1\_7\_0\_4*} & The perceived increase in extreme weather is artificial & N/A \\
\multicolumn{1}{l}{} & \multicolumn{1}{l}{1\_7\_0\_5*} & The perceived increase in extreme weather is due to improper metrics & N/A \\
\multicolumn{1}{l}{} & \multicolumn{1}{l}{1\_7\_0\_6*} & The idea that extreme weather is increasing is wrong or unsupported & N/A \\
\multicolumn{1}{l}{} & \multicolumn{1}{l}{1\_7\_0\_7*} & Extreme weather events are milder than in the past & N/A \\ \hline
\end{tabular}
\caption{Taxonomy of claims used for the climate change contrarian claims detection task. Claims in the right column are the negated claims used in Experiment 3, while those with an asterix are the extra claims used for Experiment 4. }\label{tab:clim_taxonomy}
\end{table*}

\begin{table*}[!htb]
\centering
\scriptsize
\begin{tabular}{lll}
\hline
\multicolumn{1}{c}{\textbf{Class}} & \multicolumn{1}{c}{\textbf{Claim ID}} & \multicolumn{1}{c}{\textbf{Claim}}                       \\ \hline
1 (Atheism)                                  & 1\_0                                  & This text is about atheism or religion                   \\
   & 1\_1                                  & This text is about atheism                               \\
   & 1\_2                                  & This text is about religion                              \\
   & 1\_3                                  & This text is about God                                   \\
   & 1\_4                                  & This text is about the Bible                             \\ \hline
2 (Climate change)                                  & 2\_0                                  & This text is about climate change                        \\
    & 2\_1                                  & This text is about climate action                        \\
    & 2\_2                                  & This text is about global warming                        \\
    & 2\_3                                  & This text is about the climate movement                  \\
    & 2\_4                                  & This text is about the climate change scam               \\ \hline
3 (Feminism)                                  & 3\_0                                  & This text is about feminism                              \\
   & 3\_1                                  & This text is about women's rights                        \\
   & 3\_2                                  & This text is about sexism                                \\
   & 3\_3                                  & This text is about women's privilege                     \\
   & 3\_4                                  & This text is about women's inferiority                   \\ \hline
4 (Hillary Clinton)                                  & 4\_0                                  & This text is about Hillary Clinton                       \\
   & 4\_1                                  & This text is about the Democratic presidential candidate \\
   & 4\_2                                  & This text is about the presidential election             \\
   & 4\_3                                  & This text is about Bernie Sanders                        \\
   & 4\_4                                  & This text is about Donald Trump                          \\ \hline
5 (Abortion)                                  & 5\_0                                  & This text is about abortion                              \\
   & 5\_1                                  & This text is about reproductive rights                   \\
   & 5\_2                                  & This text is about murdering babies                      \\
   & 5\_3                                  & This text is about pregnancy                             \\
   & 5\_4                                  & This text is about foetus                                \\ \hline
\end{tabular}
\caption{Taxonomy of claims used for the topic detection component of the topic/stance classification task.}\label{tab:topic_taxonomy}
\end{table*}

\begin{table*}[!htb]
\tiny
\centering
\begin{tabular}{lll|lll}
\hline
\textbf{Class}  & \textbf{Claim ID} & \textbf{Claim} & \textbf{Class}  & \textbf{Claim ID} & \textbf{Claim} \\ \hline
\multirow{7}{*}{\begin{tabular}[c]{@{}l@{}}1A \\ (Atheism, Against)\end{tabular}}  & 1A\_1  & God exists  & \multirow{7}{*}{\begin{tabular}[c]{@{}l@{}}1F \\ (Atheism, In favor)\end{tabular}}   & 1F\_1   & God does not exist \\
       & 1A\_2             & We need religion &  & 1F\_2             & Religion causes suffering   \\
       & 1A\_3             & We need God  & & 1F\_3  & Religion is a lie  \\
       & 1A\_4             & Jesus will save us &   & 1F\_4 & Atheism is the way to go \\
       & 1A\_5             & Atheism is bad &  & 1F\_5  & I am an atheist  \\
       & 1A\_6             & I believe in God  &  & 1F\_6  & I oppose religion  \\
       & 1A\_7             & I oppose atheism &  &  & \\ \hline
\multirow{6}{*}{\begin{tabular}[c]{@{}l@{}}2A\\ (Climate change, \\ Against)\end{tabular}} & 2A\_1             & Climate change is not happening                   & \multirow{6}{*}{\begin{tabular}[c]{@{}l@{}}2F\\ (Climate change, \\ In favor)\end{tabular}}  & 2F\_1             & Climate change is happening                                                                                                          \\
       & 2A\_2             & Humans are not causing climate change &  & 2F\_2  & Humans are causing climate change  \\
       & 2A\_3             & Climate change is not a problem   & & 2F\_3  & Climate change threatens our way of life \\
       & 2A\_4             & Climate change is a hoax & & 2F\_4  & Climate change is real \\
       & 2A\_5             & The climate movement is hypocritical &  & 2F\_5 & The climate movement should be taken seriously \\
       & 2A\_6             & I oppose the climate movement & & 2F\_6  & I support the climate movement \\ \hline
\multirow{6}{*}{\begin{tabular}[c]{@{}l@{}}3A\\ (Feminism, Against)\end{tabular}}          & 3A\_1             & Feminists have no humor                           & \multirow{6}{*}{\begin{tabular}[c]{@{}l@{}}3F\\ (Feminism, In favor)\end{tabular}}           & 3F\_1             & Women and men should be treated as equals \\
       & 3A\_2             & Feminism is a threat to society                   &                                                                                              & 3F\_2             & Women have the same abilities as men \\
       & 3A\_3             & Feminism is bullshit                              &                                                                                              & 3F\_3             & We should fight for women’s rights \\
       & 3A\_4             & Feminism goes against the natural order of things &                                                                                              & 3F\_4             & Men oppress women \\
       & 3A\_5             & Women have it easy                                &                                                                                              & 3F\_5             & I am a feminist \\
       & 3A\_6             & I oppose feminism                                 &                                                                                              & 3F\_6             & I support feminism  \\ \hline
\multirow{4}{*}{\begin{tabular}[c]{@{}l@{}}4A\\ (Hillary Clinton,\\ Against)\end{tabular}} & 4A\_1             & Hillary Clinton should not be president           & \multirow{4}{*}{\begin{tabular}[c]{@{}l@{}}4F\\ (Hillary Clinton, \\ In favor)\end{tabular}} & 4F\_1             & Hillary Clinton would make a good president \\
       & 4A\_2             & Hillary Clinton’s policies are bad                &                                                                                              & 4F\_2             & Hillary Clinton’s policies are good \\
       & 4A\_3             & I oppose Hillary Clinton’s candidacy              &                                                                                              & 4F\_3             & I support Hillary Clinton’s candidacy \\
       & 4A\_4             & I will not vote for Hillary Clinton               &                                                                                              & 4F\_4             & I will vote for Hillary Clinton \\ \hline
\multirow{8}{*}{\begin{tabular}[c]{@{}l@{}}5A\\ (Abortion, Against)\end{tabular}}          & 5A\_1             & Abortion is child murder                          & \multirow{8}{*}{\begin{tabular}[c]{@{}l@{}}5F\\ (Abortion, In favor)\end{tabular}}           & 5F\_1             & Abortion is not murder \\
       & 5A\_2             & Abortion should be illegal                        &                                                                                              & 5F\_2             & Access to abortion is a woman’s right \\
       & 5A\_3             & Abortion is immoral                               &                                                                                              & 5F\_3             & Access to abortion is important for women’s health  \\
       & 5A\_4             & Abortion is dangerous for women’s health          &                                                                                              & 5F\_4             & Women should have the option to get an abortion \\
       & 5A\_5             & Abortion causes suffering                         &                                                                                              & 5F\_5             & The right to abortion is good for society \\
       & 5A\_6             & Abortion clinics should be shut down              &                                                                                              & 5F\_6             & The right to abortion prevents suffering \\
       & 5A\_7             & I oppose abortion                                 &                                                                                              & 5F\_7             & \begin{tabular}[c]{@{}l@{}}Restricting access to abortion amounts to restricting \\ women’s control over their own body\end{tabular} \\
       &                   &                                                   &                                                                                              & 5F\_8             & I support the right to get an abortion                                                                                               \\ \hline
\end{tabular}
\caption{Taxonomy of claims used for the stance detection component of the topic/stance classification task.}\label{tab:stance_taxonomy}
\end{table*}

\begin{table*}[!htb]
\tiny
\centering
\begin{tabular}{lll|lll}
\hline
\multicolumn{1}{c}{\textbf{Class}}                                                    & \multicolumn{1}{c}{\textbf{Claim ID}} & \multicolumn{1}{c|}{\textbf{Claim}}                                        & \multicolumn{1}{c}{\textbf{Class}}                                                           & \multicolumn{1}{c}{\textbf{Claim ID}} & \multicolumn{1}{c}{\textbf{Claim}}                                \\ \hline
\multirow{3}{*}{\begin{tabular}[c]{@{}l@{}}1 \\ (Sadness)\end{tabular}}               & 1\_1                                  & I feel sad much of the time                                                & \multirow{3}{*}{\begin{tabular}[c]{@{}l@{}}13\\ (Indecisiveness)\end{tabular}}               & 13\_1                                 & I find it more difficult to make decisions than usual             \\
  & 1\_2                                  & I am sad all the time                                                      &                                                                                              & 13\_2                                 & I have much greater difficulty in making decisions than I used to \\
  & 1\_3                                  & I am so sad or unhappy that I can't stand it                               &                                                                                              & 13\_3                                 & I have trouble making any decisions                               \\ \hline
\multirow{3}{*}{\begin{tabular}[c]{@{}l@{}}2 \\ (Pessimism)\end{tabular}}             & 2\_1                                  & I feel more discouraged about my future than I used to                     & \multirow{3}{*}{\begin{tabular}[c]{@{}l@{}}14 \\ (Worthlessness)\end{tabular}}               & 14\_1                                 & I don't consider myself as worthwhile and useful as I used to     \\
  & 2\_2                                  & I do not expect things to work out for me                                  &                                                                                              & 14\_2                                 & I feel more worthless as compared to others                       \\
  & 2\_3                                  & I feel my future is hopeless and will only get worse                       &                                                                                              & 14\_3                                 & I feel utterly worthless                                          \\ \hline
\multirow{3}{*}{\begin{tabular}[c]{@{}l@{}}3 \\ (Past failures)\end{tabular}}         & 3\_1                                  & I have failed more than I should have                                      & \multirow{3}{*}{\begin{tabular}[c]{@{}l@{}}15\\ (Loss of energy)\end{tabular}}               & 15\_1                                 & I have less energy than I used to have                            \\
  & 3\_2                                  & As I look back, I see a lot of failures                                    &                                                                                              & 15\_2                                 & I don't have enough energy to do very much                        \\
  & 3\_3                                  & I feel I am a total failure as a person                                    &                                                                                              & 15\_3                                 & I don't have enough energy to do anything                         \\ \hline
\multirow{3}{*}{\begin{tabular}[c]{@{}l@{}}4 \\ (Loss of \\ pleasure)\end{tabular}}   & 4\_1                                  & I don't enjoy things as much as I used to                                  & \multirow{6}{*}{\begin{tabular}[c]{@{}l@{}}16 \\ (Changes in \\ sleep pattern)\end{tabular}} & 16\_1a                                & I sleep somewhat more than usual                                  \\
  & 4\_2                                  & I get very little pleasure from the things I used to enjoy                 &                                                                                              & 16\_1b                                & I sleep somewhat less than usual                                  \\
  & 4\_3                                  & I can't get any pleasure from the things I used to enjoy                   &                                                                                              & 16\_2a                                & I sleep a lot more than usual                                     \\ \cline{1-3}
\multirow{3}{*}{\begin{tabular}[c]{@{}l@{}}5 \\ (Guilty \\ feelings)\end{tabular}}    & 5\_1                                  & I feel guilty over many things I have done or should have done             &                                                                                              & 16\_2b                                & I sleep a lot less than usual                                     \\
  & 5\_2                                  & I feel quite guilty most of the time                                       &                                                                                              & 16\_3a                                & I sleep most of the day                                           \\
  & 5\_3                                  & I feel guilty all of the time                                              &                                                                                              & 16\_3b                                & I wake up 1-2 hours early and can't get back to sleep             \\ \hline
\multirow{3}{*}{\begin{tabular}[c]{@{}l@{}}6 \\ (Punishment\\ feelings)\end{tabular}} & 6\_1                                  & I feel I may be punished                                                   & \multirow{3}{*}{\begin{tabular}[c]{@{}l@{}}17 \\ (Irritability)\end{tabular}}                & 17\_1                                 & I am more irritable than usual                                    \\
  & 6\_2                                  & I expect to be punished                                                    &                                                                                              & 17\_2                                 & I am much more irritable than usual                               \\
  & 6\_3                                  & I feel I am being punished                                                 &                                                                                              & 17\_3                                 & I am irritable all the time                                       \\ \hline
\multirow{3}{*}{\begin{tabular}[c]{@{}l@{}}7 \\ (Self-dislike)\end{tabular}}          & 7\_1                                  & I have lost confidence in myself                                           & \multirow{6}{*}{\begin{tabular}[c]{@{}l@{}}18 \\ (Changes in \\ appetite)\end{tabular}}      & 18\_1a                                & My appetite is somewhat less than usual                           \\
  & 7\_2                                  & I am disappointed in myself                                                &                                                                                              & 18\_1b                                & My appetite is somewhat greater than usual                        \\
  & 7\_3                                  & I dislike myself                                                           &                                                                                              & 18\_2a                                & My appetite is much less than before                              \\ \cline{1-3}
\multirow{3}{*}{\begin{tabular}[c]{@{}l@{}}8\\ (Self-\\ criticalness)\end{tabular}}   & 8\_1                                  & I am more critical of myself than I used to be                             &                                                                                              & 18\_2b                                & My appetite is much greater than usual                            \\
  & 8\_2                                  & I criticize myself for all of my faults                                    &                                                                                              & 18\_3a                                & I have no appetite at all                                         \\
  & 8\_3                                  & I blame myself for everything bad that happens                             &                                                                                              & 18\_3b                                & I crave food all the time                                         \\ \hline
\multirow{3}{*}{\begin{tabular}[c]{@{}l@{}}9 \\ (Suicidal \\ ideation)\end{tabular}}  & 9\_1                                  & I have thoughts of killing myself, but I would not carry them out          & \multirow{3}{*}{\begin{tabular}[c]{@{}l@{}}19 \\ (Concentration\\ difficulty)\end{tabular}}  & 19\_1                                 & I can't concentrate as well as usual                              \\
  & 9\_2                                  & I would like to kill myself                                                &                                                                                              & 19\_2                                 & It's hard to keep my mind on anything for very long               \\
  & 9\_3                                  & I would kill myself if I had the chance                                    &                                                                                              & 19\_3                                 & I find I can't concentrate on anything                            \\ \hline
\multirow{3}{*}{\begin{tabular}[c]{@{}l@{}}10 \\ (Crying)\end{tabular}}               & 10\_1                                 & I cry more than I used to                                                  & \multirow{3}{*}{\begin{tabular}[c]{@{}l@{}}20 \\ (Tiredness or \\ fatigue)\end{tabular}}     & 20\_1                                 & I get more tired or fatigued more easily than usual               \\
  & 10\_2                                 & I cry over every little thing                                              &                                                                                              & 20\_2                                 & I am too tired or fatigued to do a lot of the things I used to do \\
  & 10\_3                                 & I feel like crying, but I can't                                            &                                                                                              & 20\_3                                 & I am too tired or fatigued to do most of the things I used to do  \\ \hline
\multirow{3}{*}{\begin{tabular}[c]{@{}l@{}}11\\ (Agitation)\end{tabular}}             & 11\_1                                 & I feel more restless or wound up than usual                                & \multirow{3}{*}{\begin{tabular}[c]{@{}l@{}}21 \\ (Loss of interest \\ in sex)\end{tabular}}  & 21\_1                                 & I am less interested in sex than I used to be.                    \\
  & 11\_2                                 & I am so restless or agitated, it's hard to stay still                      &                                                                                              & 21\_2                                 & I am much less interested in sex now.                             \\
  & 11\_3                                 & \begin{tabular}[c]{@{}l@{}}I am so restless or agitated that I have to keep moving or doing \\ something\end{tabular} &                                                                                              & 21\_3                                 & I have lost interest in sex completely.                           \\ \cline{1-3}
\multirow{3}{*}{\begin{tabular}[c]{@{}l@{}}12\\ (Loss of \\ interest)\end{tabular}}   & 12\_1                                 & I am less interested in other people or things than before,               &                                                                                              &                                       &                                                                   \\
  & 12\_2                                 & I have lost most of my interest in other people or things,                &                                                                                              &                                       &                                                                   \\
  & 12\_3                                 & It's hard to get interested in anything,                                  &                                                                                              &                                       &                                                                   \\ \hline
\end{tabular}
\caption{Taxonomy of claims used for the depressive symptoms detection task.}
\end{table*}

\begin{table*}[!htb]
\centering
\tiny
\begin{tabular}{lll}
\hline
\multicolumn{1}{c}{\textbf{Task}} & \multicolumn{2}{c}{\textbf{Prompt}} \\ \hline
\textbf{Climate change} & \multicolumn{2}{l}{\begin{tabular}[c]{@{}l@{}}Decide whether the text implies the claim, answering with YES or NO. For example:\\ \\ Text: Given that Earth's temperature has stabilized since 1998, you can expect it to begin cooling soon because a new ice age has begun.\\ Claim: We are heading into an ice age\\ Answer: Yes\\ \\ Text: Congress and the next Administration should open access to America's abundant reserves and let states regulate energy production within their borders.\\ Claim: Global warming is not happening\\ Answer: No\\ \\ Text: \{TEXT\}\\ Claim: \{CLAIM\}\\ Answer:\end{tabular}} \\ \hline
\textbf{Topic/Stance} & \begin{tabular}[c]{@{}l@{}}Decide whether the topic relates to the text, answering with YES or NO. \\ For example:\\ \\ Text: SO EXCITING! Meaningful climate change action is on the way! \\ Topic: Climate Change is a Real Concern\\ Answer: Yes\\ \\ Text: Blessed are the peacemakers, for they shall be called children of God. Matthew 5:9\\ Topic: Feminist Movement\\ Answer: No\\ \\ Text: \{TEXT\}\\ Topic: \{TOPIC\}\\ Answer:\end{tabular} & \begin{tabular}[c]{@{}l@{}}Decide whether the topic relates to the text, answering with ANTI, PRO or \\ NEUTRAL. For example:\\ \\ Text: SO EXCITING! Meaningful climate change action is on the way! \\ Topic: Climate Change is a Real Concern\\ Stance: Pro\\ \\ Text: Let's agree that it's not ok to kill a 7lbs baby in the uterus\\ Topic: Legalization of Abortion\\ Stance: Anti\\ \\ Text: \{TEXT\}\\ Topic: \{TOPIC\}\\ Stance:\end{tabular} \\ \hline
\textbf{Depression} & \multicolumn{2}{l}{\begin{tabular}[c]{@{}l@{}}Decide whether the text implies the claim, answering with YES or NO. For example:\\ \\ Text: Everything I used to be passionate about has gone down the drain.\\ Claim: It's hard to get interested in anything\\ Answer: Yes\\ \\ Text: I cry and care too much, which leaves me burnt out and exhausted.\\ Claim: I dislike myself\\ Answer: No\\ \\ Text: \{TEXT\}\\ Claim: \{CLAIM\}\\ Answer:\end{tabular}} \\ \hline
\end{tabular}
\caption{Prompts used for each task for the Llama 2 baseline.}\label{tab:prompt_examples}
\end{table*}

\end{document}